\documentclass[conference]{IEEEtran}
\IEEEoverridecommandlockouts
\usepackage{cite}
\usepackage{amsmath}
\usepackage{algorithmic}
\usepackage{graphicx}
\usepackage{caption}
\usepackage{subcaption}
\usepackage{textcomp}
\usepackage{xcolor}
\usepackage{nicefrac}
\usepackage{bm}
\usepackage{moresize}

\def\BibTeX{{\rm B\kern-.05em{\sc i\kern-.025em b}\kern-.08em
    T\kern-.1667em\lower.7ex\hbox{E}\kern-.125emX}}

\begin{document}
\onecolumn 

{\Huge \bfseries  IEEE Copyright Notice}

\vspace{20mm}

\large ©2023 IEEE. Personal use of this material is permitted. Permission from IEEE must be obtained for all other uses, in any current or future media, including reprinting/republishing this material for advertising or promotional purposes, creating new collective works, for resale or redistribution to servers or lists, or reuse of any copyrighted
component of this work in other works.

\vspace{10mm}

\normalsize \textit{Accepted for publication, 57th Asilomar Conference on Signals, Systems and Computers, October, 2023, Pacific Grove, California, USA.}

\newpage

\twocolumn

\title{Trust, but Verify: Robust Image Segmentation using Deep Learning}

\author{Fahim Ahmed Zaman, Xiaodong Wu, Weiyu Xu, Milan Sonka and Raghuraman Mudumbai
\thanks{Fahim Ahmed Zaman, Xiaodong Wu, Weiyu Xu, Milan Sonka and Raghu Mudumbai are with the Department of Electrical and Computer Engineering, University of Iowa, Iowa City, IA 52242, USA (e-mails: \{fahim-zaman, xiaodong-wu, weiyu-xu, milan-sonka, raghuraman-mudumbai\}@uiowa.edu}
\thanks{This research was supported in part by NIH NIBIB Grant R01-EB004640.}}

\maketitle

\begin{abstract}
We describe a method for verifying the output of a deep neural network for medical image segmentation that is robust to several classes of random as well as worst-case perturbations i.e. adversarial attacks. This method is based on a general approach recently developed by the authors called ``Trust, but Verify" wherein an auxiliary verification network produces predictions about certain masked features in the input image using the segmentation as an input. A well-designed auxiliary network will produce high-quality predictions when the input segmentations are accurate, but will produce low-quality predictions when the segmentations are incorrect. Checking the predictions of such a network with the original image allows us to detect bad segmentations. However, to ensure the verification method is truly robust, we need a method for checking the quality of the predictions that does not itself rely on a black-box neural network. Indeed, we show that previous methods for segmentation evaluation that do use deep neural regression networks are vulnerable to false negatives i.e. can inaccurately label bad segmentations as good. We describe the design of a verification network that avoids such vulnerability and present results to demonstrate its robustness compared to previous methods. \end{abstract}

\begin{IEEEkeywords}
robust deep learning, adversarial attack, image segmentation
\end{IEEEkeywords}

\section{Introduction}

Recent works, including our own \cite{Hesamian, xie2022, Zhang2021, peng2022, Zhao2018}, have shown that deep neural networks (DNN) can deliver significant improvements in medical image segmentation. However, the black-box character of DNNs makes it challenging to precisely identify their capabilities and limitations. In particular, it is difficult to know if their outputs can be trusted to be correct for a given set of inputs. This is of course a well-known challenge in machine learning that has attracted enormous ongoing interest among researchers.

We have developed a general approach to this problem loosely described by the phrase ``Trust, but Verify". The idea is to augment the neural-network based segmentation algorithm with an auxiliary generative network that will use the output of the segmentation network to ``paint in" certain masked features of the original image. A well-designed verification network will produce high-quality prediction when the input segmentation is accurate. In case of an incorrect segmentation, the prediction of the verification network will be of low-quality. Checking the predictions of such a network with the original image allows us to differentiate between correct and incorrect segmentations.

Note that a conservative design is appropriate for applications such as medical image segmentation where mistakes can have serious consequences. This means that the verification network must be designed to minimize Type II errors i.e. false negatives where bad segmentations are incorrectly classified as good. Type I errors i.e. false positives where good segmentations are incorrectly rejected, are not as serious as Type II errors. A small rate of false positives may be acceptable, but a good verification network should effectively rule out the possibility of false negatives. In other words, with a good verification network it should be virtually impossible for a bad segmentation to be classified as good.

\subsection{Problem Statement: Robust Segmentation Evaluation}

The main motivation for the present work is that previous tools for evaluating the quality of image segmentations do not meet this criterion. Specifically, we present evidence to show that previous methods for segmentation evaluation are vulnerable to false negatives i.e. can inaccurately evaluate bad segmentations as good.

Our ``Trust, but Verify" approach is not a magic bullet solution to determine the correctness of neural network outputs. In particular, designing a well-functioning verification network presents a considerable challenge that needs to be addressed in an application-specific manner. In this paper, we present one possible design for a verification network for medical image segmentation. We present experimental results to show its effectiveness and also its superior robustness compared to state-of-the-art designs.

\subsection{Previous Work on Segmentation Evaluation} \label{sec:survey}

Traditional methods for segmentation evaluation such as region-based, boundary-based and hybrid measures \cite{Feng,Huttenlocher,Movahedi} greatly rely on the use of available reference segmentations. In the absence of case-by-case reference segmentations for large 3D medical image datasets, the problem of segmentation-quality evaluation remains largely open. Recently, deep learning based methods have been proposed, which learn the inherent image features for segmentation evaluation \cite{Zhong, weifan, Kohlberger, CIBM, JMI}. Huang et al. \cite{Huang} proposed three independent deep neural networks for segmentation evaluation. The first two use CNNs and FCNs (Full Convolutional Networks) for foreground mask prediction to evaluate the segmentation quality by predicting Dice Similarity Co-efficient (DSC). The third network uses weighted mask layers to extract mask features for foreground and background information analysis to predict DSC. Zhou et al. \cite{zhou} proposed to use a reconstruction network for robust feature extraction that is directly related to the segmentation and to utilize a regression network to predict DSC scores from the extracted features. Robinson et al. \cite{robinson} proposed to use a Convolutional Neural Network (CNN) based regression network by directly feeding the image and its segmentation as different channels for DSC prediction. All these approaches aim to evaluate the quality of segmentation by computing a single overall accuracy score.

\begin{figure}[h]
\centering
    \includegraphics[width=0.45\textwidth]{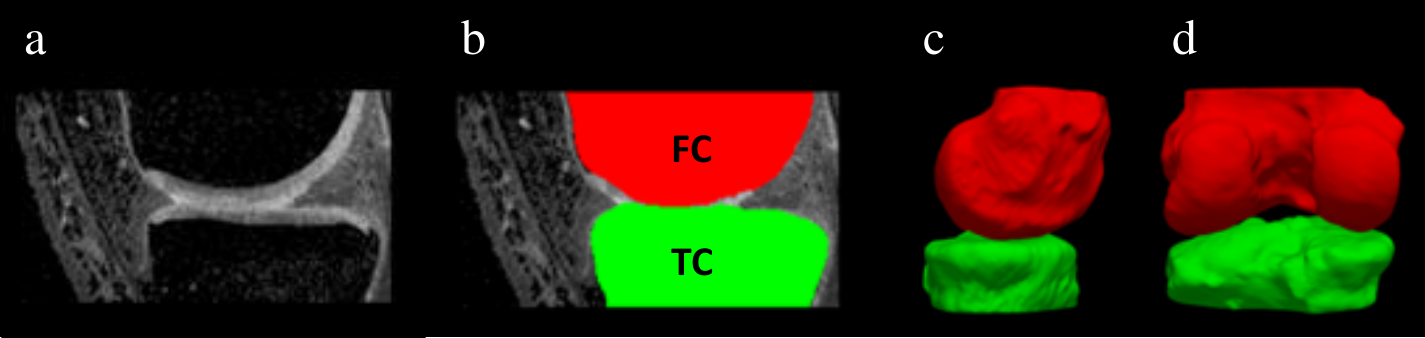}
    \caption{3D Knee-MR dataset. (a) A sample slice of knee-MR dataset in sagittal plane, (b) Reference segmentation masks FC (red) and TC (green), (c-d) 3-D surfaces of FC and TC in the sagittal and coronal views, respectively.}
\label{fig:dataset} 
\end{figure}

\subsection{Dataset}

We used Osteoarthritis Initiative (OAI) 3D knee MRI (knee-MR) (https://data-archive.nimh.nih.gov/oai/) for method development and evaluation. The OAI 3D knee MRI dataset consists of 1462 double echo steady state (DESS) 3D MR images from 248 subjects. The image size is $384\times384\times160$, with voxel size of $0.36mm\times0.36mm\times0.70mm$. The dataset is first segmented by the LOGISMOS method \cite{Kashyap}, and then corrected by the Just-Enough-Interaction (JEI) approach \cite{JEI}. Four compartments are annotated: femur bone (FB), femoral cartilage (FC), tibia bone (TB), and tibial cartilage (TC). FC and TC surfaces produced by LOGISMOS are the combination of outer surface of cartilage and the bone compartment. We only used two annotated compartments, FC and TC, in our experiments. We focus on the contact region by selecting ROIs having the size of $104\times240\times160$. The middle slice of the segmentation results of FC and TC is first calculated. Here, middle slice is the slice in the transverse plane, which has the equal distance from the outermost surface voxels of FC and TC. A bounding box of size $104\times240\times160$ is then put around the center of the middle slice to extract the ROI. A sample ROI of knee MRI slice along with its FC and TC segmentation maps and 3D surfaces are shown in Fig. \ref{fig:dataset}

\begin{figure}[h]
\centering
    \includegraphics[width=0.45\textwidth]{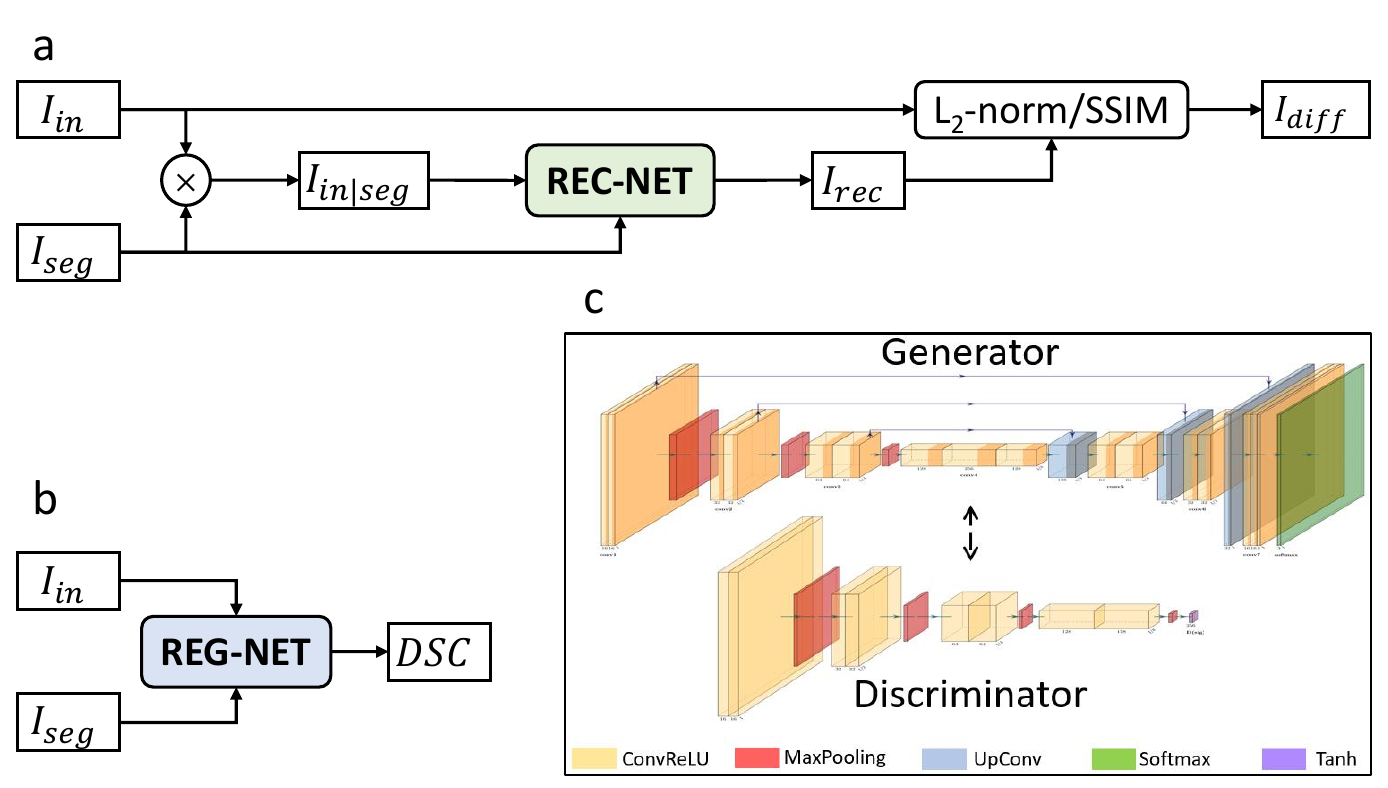}
    \caption{(a) Method workflow for proposed conditional GAN based REC-Net, (b) Method workflow for REG-Net, (c) Generator and Discriminator architecture for REC-Net. The discriminator is only used in the training phase of the model.}
\label{fig:method} 
\end{figure}

\subsection{Failure of Robustness}

We implemented a fully convolutional regression network following the method proposed by Robinson et al. \cite{robinson} for segmentation quality assessment. Our implementation of REG-Net is a state-of-the-art regression network which is able to produce extremely accurate predictions of the DSC. The method workflow for REG-Net is shown in Fig. \ref{fig:method}b; the network takes the original image and its corresponding predicted segmentation mask as two channel inputs and outputs a predicted DSC score for the given image-segmentation pair.  

\begin{figure}[h]
\centering
    \includegraphics[width=0.25\textwidth]{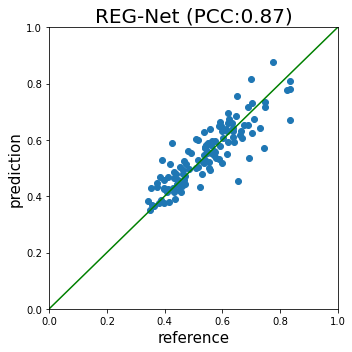}
    \caption{Predicted DSC score from REG-Net against reference DSC scores.}
\label{fig:regdice} 
\end{figure}

Fig. \ref{fig:regdice} plots the predicted DSC score against reference segmentations manually created by imaging experts. It is clear that the network provides accurate predictions over a wide range of DSCs. The Pearson correlation co-efficient (PCC) of REG-Net for the knee-MR dataset is $87\%$. 

Unfortunately, the regression network, being a black-box deep neural network, itself represents a potential point of failure. In particular, there is a question of whether the REG-Net can be ``fooled" into outputting a high quality score even with bad segmentations. As noted earlier, it is essential for a good verification network to provide robust safeguards against such Type II errors. 

\begin{figure}[h]
\centering
    \includegraphics[width=0.4\textwidth]{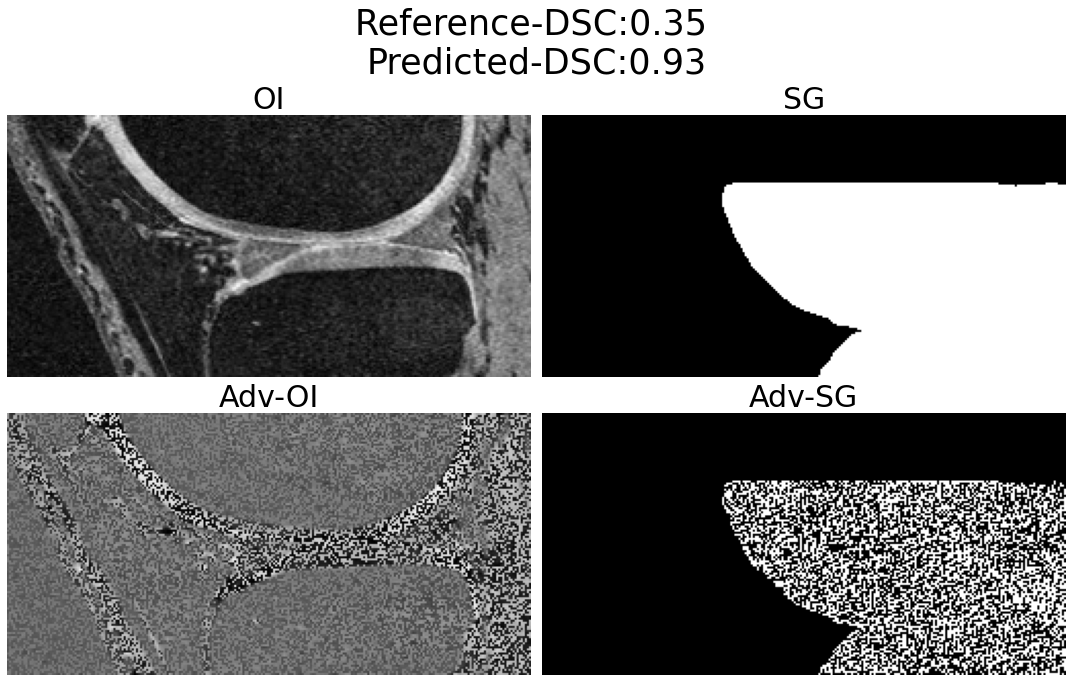}
    \caption{Original image (OI) and predicted incorrect segmentation (SG) are shown in top row and their respective perturbed images with FGSM using $\epsilon=0.5$ are shown in the bottom row.}
\label{fig:regnetadv} 
\end{figure}

We investigated this question by attempting to construct adversarial examples for the REG-Net. It turns out that such examples can be readily constructed using standard methods. A proof-of-concept example image is shown in Fig. \ref{fig:regnetadv} where REG-Net outputs a predicted DSC score as high as $93\%$, whereas the reference DSC score is only $35\%$. Details about the construction of the example in Fig. \ref{fig:regnetadv} are in Section \ref{sec:results}.

\section{Proposed Method}

Clearly we need a more robust replacement for the regression network REG-Net. We describe now an initial design and refinement of a verification network for robust segmentation evaluation.

\subsection{Version 1: A Generative Verification Network}

Our first attempt at an alternative verification network is shown in Fig. \ref{fig:method}a. It includes a generative network (REC-Net) that takes original image $I_{in}$ masked by a strip around the boundary of the segmentation $I_{seg}$ and the mask itself as two channel inputs. The generator is a U-Net like auto-encoder and the discriminator is a VGG like regression network. The original image and its corresponding mask are split into several independent smaller patches maintaining no overlapping among them and fed into the REC-Net. The reconstructed patches are stitched together maintaining their spatial location to recreate the final reconstructed image $I_{rec}$ that has the same size of $I_{in}$. The patch-wise reconstruction is adapted to remove the location specific bias in training the generator. The goal of REC-Net is to reconstruct the original image from the masked one $I_{in|I_{seg}}$. The architecture of the generator and the discriminator are shown in Fig. \ref{fig:method}c. Details about the generator and discriminator architectures can be found in our previous work \cite{JMI}. 

\begin{figure*}[ht!]
\centering
    \includegraphics[width=0.7\textwidth]{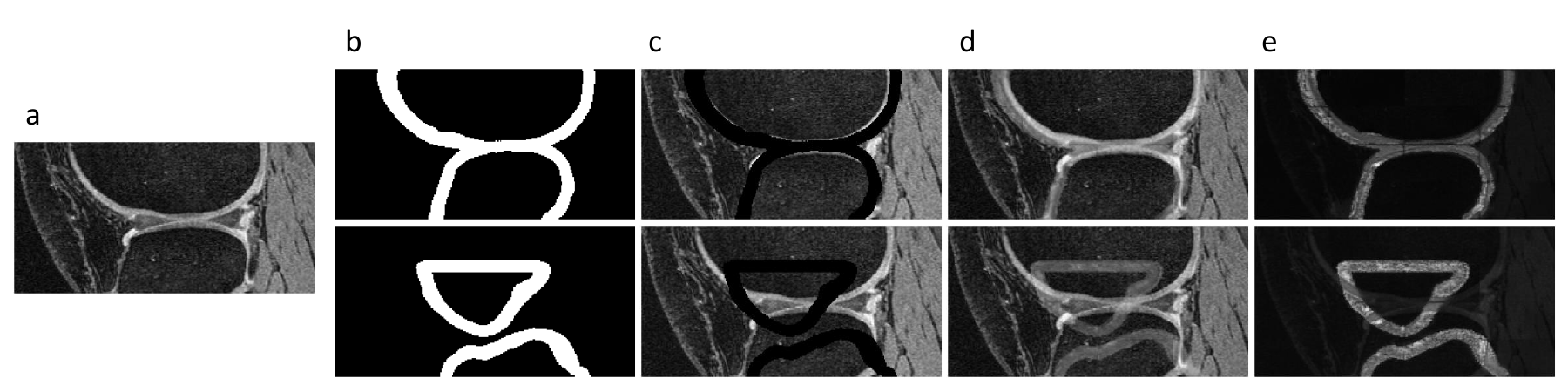}
    \caption{Reconstruction results of REC-Net in case of correct and incorrect segmentations. The top and bottom row shows the input images, reconstructions and difference images for correct and incorrect segmentations, respectively. (a) A sample knee-MR slice $I_{in}$, (b) Boundary of the predicted segmentation $I_{seg}$, (c) Masked original image with segmentation boundary, $I_{in|I_{seg}}$, (d) Reconstructed image $I_{rec}$, (e) Difference image $I_{diff}$}
\label{fig:recnetcomp}
\end{figure*}

The REC-Net is trained only with the reference segmentations. The rationale is that REC-Net is trained to well recover the original image from the masked one only when the mask segmentation $I_{seg}$ is close to the reference segmentation, whereas the REC-Net fails to produce good reconstruction in case of incorrect segmentation in the inference phase. The idea is to check if the difference $I_{diff}$ between the original and the reconstructed image is small. We have shown that REC-Net is effective at doing this as illustrated in Fig. \ref{fig:recnetcomp}.

We adopted the same strategy to train the REC-Net as mentioned in \cite{pix2pix}, where the discriminator is trained with cross-entropy loss and the generator is trained with a weighted combination of discriminator and mean absolute error (MAE) loss.

It is visually clear from Fig. \ref{fig:recnetcomp} that the quality of the predicted image from REC-Net can be a good candidate to identify bad segmentations. To test out this hypothesis, we need a non-visual {\it algorithmic} method to evaluate the quality of the predicted image from REC-Net. Furthermore, to avoid replicating the robustness issues of REG-Net, we seek a method to score the quality of the predicted image from REC-Net that does not require another black-box neural network. 

The REC-Net essentially reduces the problem of {\it segmentation evaluation} to the potentially simpler problem of {\it image prediction} evaluation. However, this is still a non-trivial problem to solve algorithmically: a naive method of simply looking at the size (e.g. $L_2$ norm) of the pixel prediction error turns out to be ineffective.

\subsection{Version 2: An Improved Verification Network}

After some experimentation, we found that the structural similarity index measure (SSIM) \cite{wang2004} is an effective metric of image quality that can be calculated algorithmically and is able to effectively discriminate between image predictions resulting from correct and incorrect segmentations. However, this requires incorporating the SSIM into the loss function used for training the REC-Net instead of MAE. 

The structural similarity (SSIM) index provides a measure of the similarity by comparing two images based on luminance similarity, contrast similarity and structural similarity information. Luminance of two image patch $x$ and $y$ can be compared by,

\begin{equation}
    l(x,y) = \frac{2\mu_{x}\mu_{y}+C_{1}}{\mu_{x}^{2}+\mu_{y}^{2}+C_{1}},
    \label{luminance}
\end{equation}
where $\mu_{x}$ and $\mu_{y}$ are the mean image intensity of $x$ and $y$. Contrast similarity of $x$ and $y$ can be compared by,

\begin{equation}
    c(x,y) = \frac{2\sigma_{x}\sigma_{y}+C_{2}}{\sigma_{x}^{2}+\sigma_{y}^{2}+C_{2}},
    \label{contrast}
\end{equation}
where $\sigma_{x}$ and $\sigma_{y}$ are the standard deviation of $x$ and $y$. Structural similarity of $x$ and $y$ can be compared as,

\begin{equation}
    s(x,y) = \frac{\sigma_{xy}+C_{3}}{\sigma_{x}\sigma_{y}+C_{3}},
    \label{structure}
\end{equation}
where $\sigma_{xy}$ is the co-variance matrix of $x$ and $y$. The constants C1, C2 and C3 are used to avoid divide-by-zero errors. Now, the SSIM of $x$ and $y$ can be constructed as,

\begin{equation}
    SSIM(x,y) = l(x,y) \times c(x,y) \times s(x,y)
    \label{ssimscore}
\end{equation}

\begin{figure}[ht]
\centering
    \includegraphics[width=0.45\textwidth]{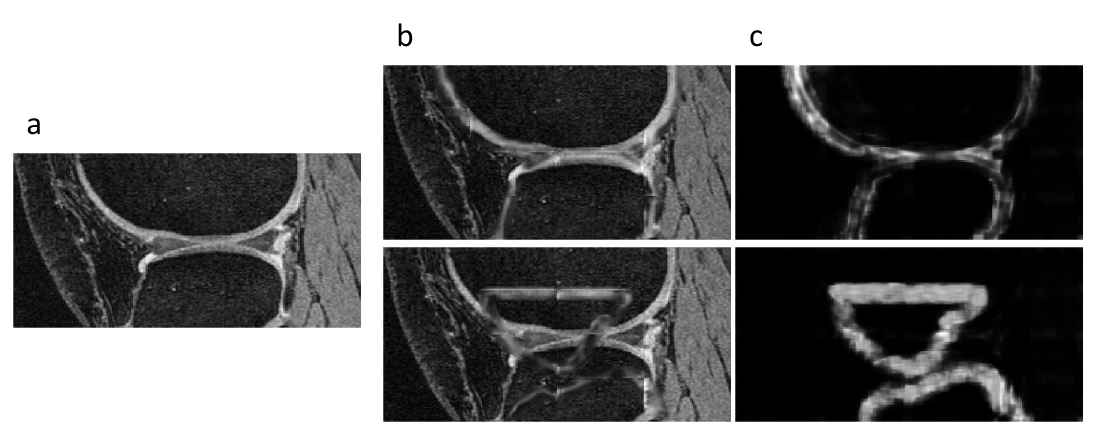}
    \caption{Reconstruction results of REC-Net trained with SSIM loss in case of correct and incorrect segmentations. The top and bottom row shows the reconstructions and difference images for correct and incorrect segmentations, respectively. (a) A sample knee-MR slice $I_{in}$, (b) Reconstructed image $I_{rec}$, (c) Difference image based on structural dissimilarity $I_{diff}$}
\label{fig:recnetssim}
\end{figure}

SSIM is calculated as a local measure rather than a global measure in order to incorporate the fact that the human visual system (HVS) can perceive only a local area at high resolution at a time. Here, $x$ and $y$ are image patches of the full images. We incorporate SSIM loss as reconstruction loss to train the generator replacing the MAE loss as,

\begin{equation}
    \mathcal{L}_{SSIM} = 1 - SSIM(I_{rec}^{*},I_{gt}^{*}),
    \label{ssimloss}
\end{equation}
whereas rest of the loss function is unchanged from \cite{pix2pix}. $I_{rec}^{*}$ and $I_{gt}^{*}$ are the image patches of $I_{rec}$ and $I_{gt}$. The reconstruction results from the updated REC-Net trained with SSIM loss instead of MAE is shown in Fig. \ref{fig:recnetssim}.

\section{Experimental Results} \label{sec:results}

We now present a description of our experimental procedures and summarize the results.

\subsection{Adversarial Examples for REG-Net}

We used the simple fast gradient signed method (FGSM) \cite{KurakinGB16} to generate adversarial images for the REG-Net, one of which is shown in Fig. \ref{fig:regnetadv}. Assume $X$ is an input vector consisting of original gray-scale image $I_{in}$ and predicted segmentation $I_{seg}$, $y$ is its reference DSC score and $J(X,y)$ is the mean squared error (MSE) cost function of the REG-Net. Then the gray-scale image and predicted segmentation with adversarial noises are generated as:
\begin{equation}
    X^{*} = X + \epsilon \cdot sign (\nabla_{X}J(X,y)) \;\; 
    \label{adversarial_eq}
\end{equation}
where $\epsilon$ is a hyper-parameter that determines the magnitude of the adversarial noise; larger the value of $\epsilon$, higher the adversarial perturbations. As noted earlier, REG-Net outputs a high predicted DSC score of $93\%$ for this clearly anomalous input.

\subsection{Segmentation Evaluation using Generative Verification Network}

\begin{figure}
     \centering
     \begin{subfigure}[a]{0.3\textwidth}
         \centering
         \includegraphics[width=\textwidth]{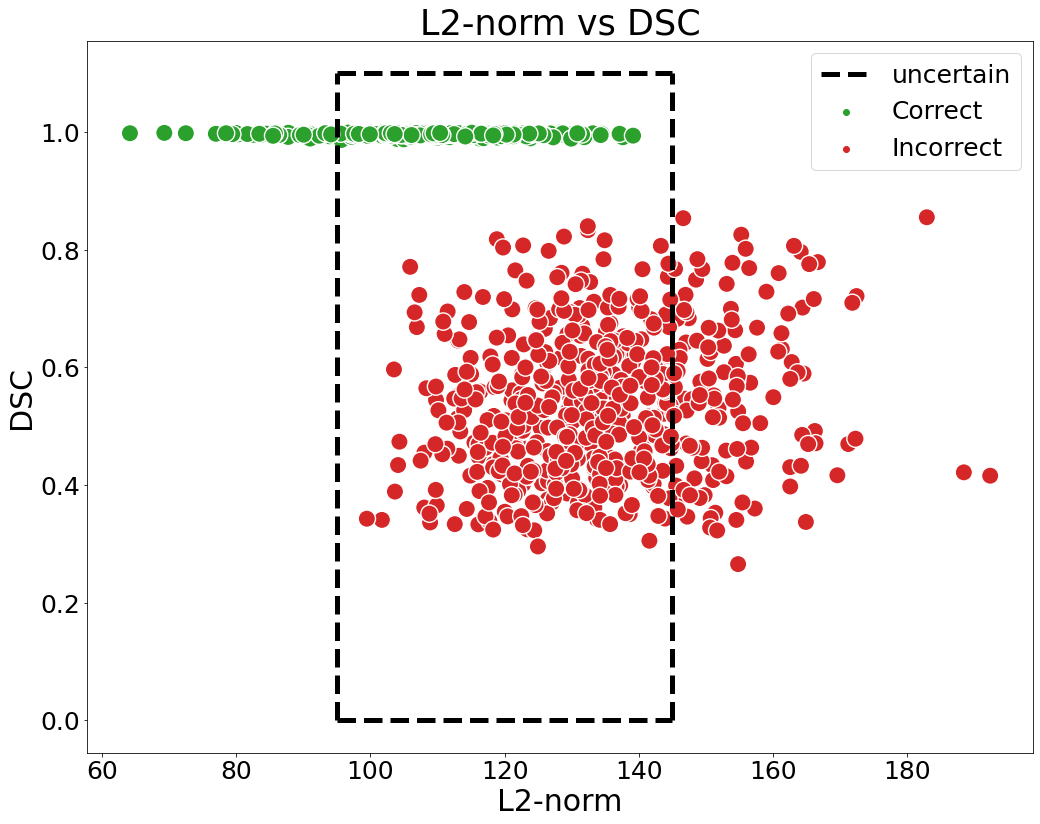}
         \caption{}
         \label{fig:l2dice}
     \end{subfigure}
     \hfill
     \begin{subfigure}[b]{0.3\textwidth}
         \centering
         \includegraphics[width=\textwidth]{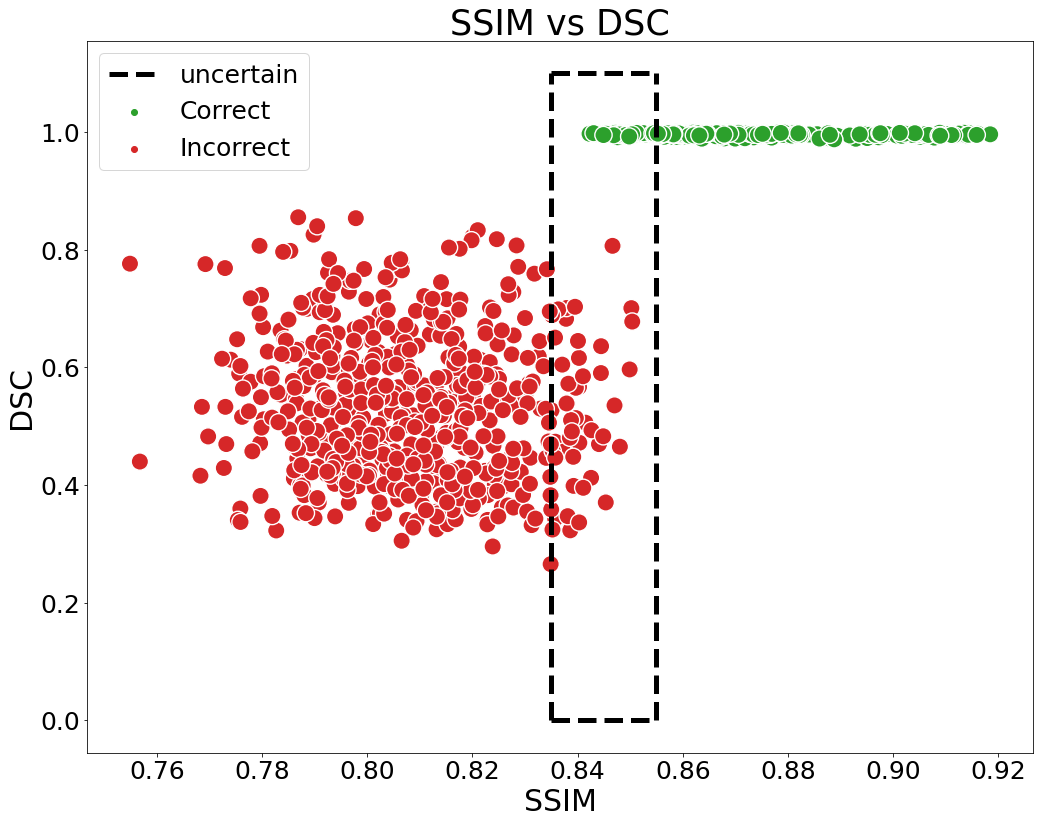}
         \caption{}
         \label{fig:ssimdice}
     \end{subfigure}
        \caption{(a) $L_{2}$ norm vs. reference DSC score, (b) SSIM vs. reference DSC score.}
        \label{fig:performance}
\end{figure}

Fig. \ref{fig:l2dice} shows the $L_2$ prediction error of the original version 1 of the REC-Net against the reference DSC score for correct and incorrect segmentations. We can see that on average good segmentations have smaller prediction errors. However, the size of the $L_2$ errors is not a reliable way to differentiate between good and bad segmentations.

By contrast, consider Fig. \ref{fig:ssimdice} which plots the SSIM of the improved version 2 of the REC-Net against the reference DSC score. While the discrimination between correct and incorrect segmentations is not perfect as seen from the ``uncertain region" in Fig. \ref{fig:ssimdice}, it is clear that Fig. \ref{fig:ssimdice} represents a vast improvement over Fig. \ref{fig:l2dice}. In particular, by applying a simple threshold on the SSIM we can effectively detect all bad segmentations while rejecting relatively few good segmentations. Further refinement of the REC-Net to improve on these preliminary results is a goal for future work.

\begin{figure}[ht]
\centering
    \includegraphics[width=0.4\textwidth]{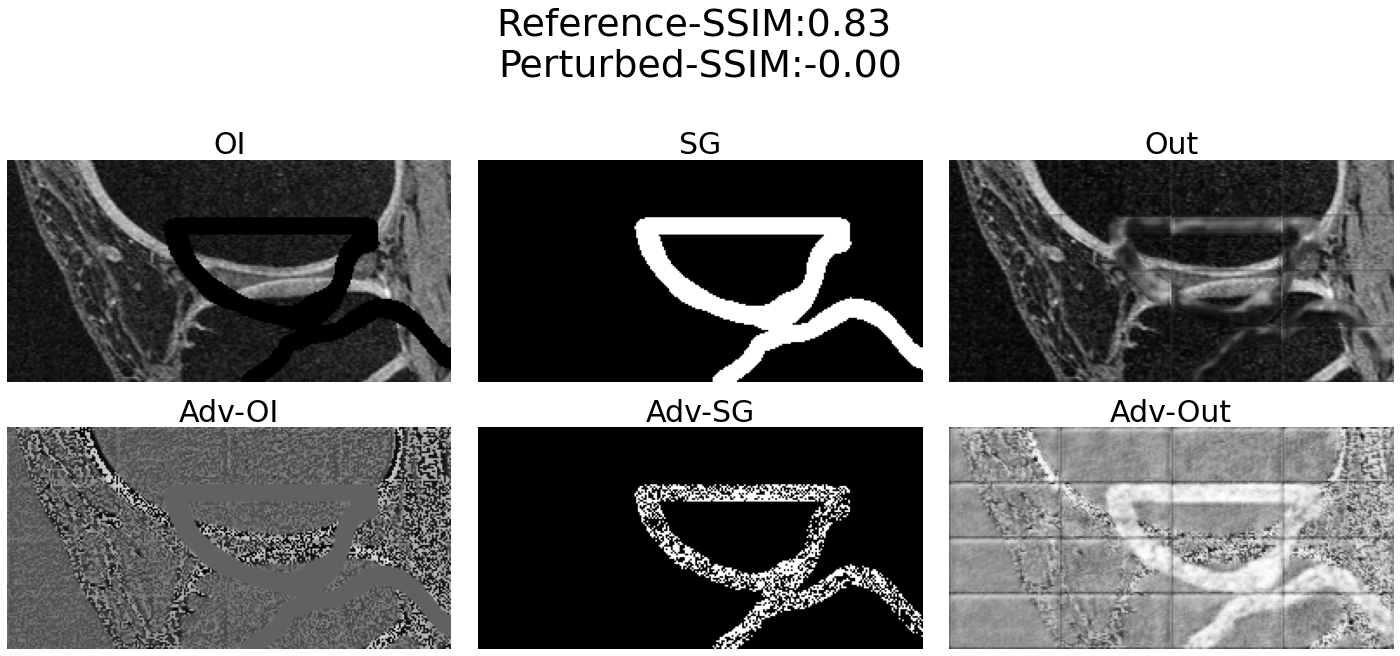}
    \caption{Original image (OI), predicted incorrect segmentation boundary (SG) and reconstruction in top row and FGSM perturbed images and reconstruction in bottom row.}
\label{fig:recAdv}
\end{figure}

Importantly, our attempts to create an adversarial example of an incorrect segmentation that can produce high SSIM scores have been completely unsuccessful. An attempted adversarial example is shown in Fig. \ref{fig:recAdv} where the perturbed image was generated using FGSM using $\epsilon=0.5$. In this and every other attempt we tried to manipulate the predicted image from REC-Net, the SSIM score only gets worse. {\it This confirms our intuition that the structural similarity index SSIM, being a well-defined mathematical construct, cannot be fooled in the same way as a black box regression network.}

\section{Conclusion}

We considered the problem of designing a robust verification method for segmentation evaluation that can allow us to safely deploy powerful but potentially unreliable segmentation algorithms that make use of black-box deep neural networks. Crucial to this ``Trust, but Verify" approach is a method that is guaranteed to detect all bad segmentations without Type II errors i.e. false negatives. We show that previous state-of-the-art methods that themselves rely on deep neural regression networks, are vulnerable to false negatives. We show that an alternative approach using a generative network optimized for a robust mathematical metric i.e. the SSIM can essentially eliminate such failures. This motivates further study of robust verification methods for other applications that can benefit from powerful, but imperfect deep neural networks.

\section*{Acknowledgments}
The OAI is a public-private partnership comprised of five contracts (N01-AR-2-2258; N01-AR-2-2259; N01-AR-2-2260; N01-AR-2-2261; N01-AR-2-2262) funded by the National Institutes of Health, a branch of the Department of Health and Human Services, and conducted by the OAI Study Investigators. Private funding partners include Merck Research Laboratories; Novartis Pharmaceuticals Corporation, GlaxoSmithKline; and Pfizer, Inc. Private sector funding for the OAI is managed by the Foundation for the National Institutes of Health. This manuscript was prepared using an OAI public use data set and does not necessarily reflect the opinions or views of the OAI investigators, the NIH, or the private funding partners.

\bibliographystyle{IEEEtran}
\bibliography{main.bib}

\begin{thebibliography}{10}
\providecommand{\url}[1]{#1}
\csname url@samestyle\endcsname
\providecommand{\newblock}{\relax}
\providecommand{\bibinfo}[2]{#2}
\providecommand{\BIBentrySTDinterwordspacing}{\spaceskip=0pt\relax}
\providecommand{\BIBentryALTinterwordstretchfactor}{4}
\providecommand{\BIBentryALTinterwordspacing}{\spaceskip=\fontdimen2\font plus
\BIBentryALTinterwordstretchfactor\fontdimen3\font minus
  \fontdimen4\font\relax}
\providecommand{\BIBforeignlanguage}[2]{{%
\expandafter\ifx\csname l@#1\endcsname\relax
\typeout{** WARNING: IEEEtran.bst: No hyphenation pattern has been}%
\typeout{** loaded for the language `#1'. Using the pattern for}%
\typeout{** the default language instead.}%
\else
\language=\csname l@#1\endcsname
\fi
#2}}
\providecommand{\BIBdecl}{\relax}
\BIBdecl

\bibitem{Hesamian}
M.~H. Hesamian, W.~Jia, X.~He, and P.~Kennedy, ``Deep learning techniques for
  medical image segmentation: Achievements and challenges,'' \emph{Journal of
  Digital Imaging}, vol.~32, 05 2019.

\bibitem{xie2022}
H.~Xie, Z.~Pan, L.~Zhou, F.~A. Zaman, D.~Z. Chen, J.~B. Jonas, W.~Xu, Y.~X.
  Wang, and X.~Wu, ``Globally optimal oct surface segmentation using a
  constrained ipm optimization,'' \emph{Optics Express}, vol.~30, no.~2, pp.
  2453--2471, 2022.

\bibitem{Zhang2021}
L.~Zhang, Z.~Guo, H.~Zhang, E.~van~der Plas, T.~R. Koscik, P.~C. Nopoulos, and
  M.~Sonka, ``Assisted annotation in deep logismos: Simultaneous
  multi-compartment 3d mri segmentation of calf muscles,'' \emph{Medical
  physics.}, 2 2023.

\bibitem{peng2022}
Y.~Peng, H.~Zheng, P.~Liang, L.~Zhang, F.~Zaman, X.~Wu, M.~Sonka, and D.~Z.
  Chen, ``Kcb-net: A 3d knee cartilage and bone segmentation network via sparse
  annotation,'' \emph{Medical image analysis}, vol.~82, p. 102574, 2022.

\bibitem{Zhao2018}
X.~Zhao, Y.~Wu, G.~Song, Z.~Li, Y.~Zhang, and Y.~Fan, ``A deep learning model
  integrating fcnns and crfs for brain tumor segmentation,'' \emph{Medical
  Image Analysis}, vol.~43, pp. 98 -- 111, 2018.

\bibitem{Feng}
\BIBentryALTinterwordspacing
S.~W. Feng~Ge and T.~Liu, ``New benchmark for image segmentation evaluation,''
  \emph{J. Electronic Imaging}, vol.~16, no.~3, p. 033011, 2007. [Online].
  Available: \url{https://doi.org/10.1117/1.2762250}
\BIBentrySTDinterwordspacing

\bibitem{Huttenlocher}
D.~Huttenlocher, G.~Klanderman, and W.~Rucklidge, ``Comparing images using the
  hausdorff distance,'' \emph{IEEE Transactions on Pattern Analysis and Machine
  Intelligence}, vol.~15, no.~9, pp. 850--863, 1993.

\bibitem{Movahedi}
V.~Movahedi and J.~H. Elder, ``Design and perceptual validation of performance
  measures for salient object segmentation,'' in \emph{2010 IEEE Computer
  Society Conference on Computer Vision and Pattern Recognition - Workshops},
  2010, pp. 49--56.

\bibitem{Zhong}
E.~Zhong, W.~Fan, Q.~Yang, O.~Verscheure, and J.~Ren, ``Cross validation
  framework to choose amongst models and datasets for transfer learning,'' in
  \emph{Machine Learning and Knowledge Discovery in Databases}, J.~L.
  Balc{\'a}zar, F.~Bonchi, A.~Gionis, and M.~Sebag, Eds.\hskip 1em plus 0.5em
  minus 0.4em\relax Berlin, Heidelberg: Springer Berlin Heidelberg, 2010, pp.
  547--562.

\bibitem{weifan}
\BIBentryALTinterwordspacing
W.~Fan and I.~Davidson, ``Reverse testing: An efficient framework to select
  amongst classifiers under sample selection bias,'' in \emph{Proceedings of
  the 12th ACM SIGKDD International Conference on Knowledge Discovery and Data
  Mining}, ser. KDD '06.\hskip 1em plus 0.5em minus 0.4em\relax New York, NY,
  USA: Association for Computing Machinery, 2006, p. 147–156. [Online].
  Available: \url{https://doi.org/10.1145/1150402.1150422}
\BIBentrySTDinterwordspacing

\bibitem{Kohlberger}
\BIBentryALTinterwordspacing
T.~Kohlberger, V.~Singh, C.~Alvino, C.~Bahlmann, and L.~Grady, ``Evaluating
  segmentation error without ground truth,'' in \emph{Proceedings of the 15th
  International Conference on Medical Image Computing and Computer-Assisted
  Intervention - Volume Part I}, ser. MICCAI'12.\hskip 1em plus 0.5em minus
  0.4em\relax Berlin, Heidelberg: Springer-Verlag, 2012, p. 528–536.
  [Online]. Available: \url{https://doi.org/10.1007/978-3-642-33415/3\_65}
\BIBentrySTDinterwordspacing

\bibitem{CIBM}
F.~A. Zaman, L.~Zhang, H.~Zhang, M.~Sonka, and X.~Wu, ``Segmentation quality
  assessment by automated detection of erroneous surface regions in medical
  images,'' \emph{Computers in Biology and Medicine}, vol. 164, p. 107324,
  2023.

\bibitem{JMI}
F.~A. Zaman, T.~K. Roy, M.~Sonka, and X.~Wu, ``{Patch-wise 3D segmentation
  quality assessment combining reconstruction and regression networks},''
  \emph{Journal of Medical Imaging}, vol.~10, no.~5, p. 054002, 2023.

\bibitem{Huang}
C.~Huang, Q.~Wu, and F.~Meng, ``Qualitynet: Segmentation quality evaluation
  with deep convolutional networks,'' \emph{2016 Visual Communications and
  Image Processing (VCIP)}, pp. 1--4, 2016.

\bibitem{zhou}
\BIBentryALTinterwordspacing
L.~Zhou, W.~Deng, and X.~Wu, ``Robust image segmentation quality assessment,''
  2019. [Online]. Available: \url{https://arxiv.org/abs/1903.08773}
\BIBentrySTDinterwordspacing

\bibitem{robinson}
R.~Robinson and et~al., ``Subject-level prediction of segmentation failure
  using real-time convolutional neural nets,'' in \emph{1st Conference on
  Medical Imaging with Deep Learning}, ser. MIDL 2018, 2018.

\bibitem{Kashyap}
S.~Kashyap, H.~Zhang, K.~Rao, and M.~Sonka, ``Learning-based cost functions for
  {3-D and 4-D} multi-surface multi-object segmentation of knee {MRI}: Data
  from the osteoarthritis initiative,'' \emph{IEEE Transactions on Medical
  Imaging}, vol.~37, no.~5, pp. 1103--1113, 2017.

\bibitem{JEI}
S.~Kashyap, H.~Zhang, and M.~Sonka, ``Just-enough interaction approach to knee
  {MRI} segmentation: Data from the osteoarthritis initiative,'' \emph{MICCAI -
  IMIC Workshop}, 10 2016.

\bibitem{pix2pix}
P.~Isola, J.-Y. Zhu, T.~Zhou, and A.~A. Efros, ``Image-to-image translation
  with conditional adversarial networks,'' 2018.

\bibitem{wang2004}
Z.~Wang, A.~C. Bovik, H.~R. Sheikh, and E.~P. Simoncelli, ``Image quality
  assessment: from error visibility to structural similarity,'' \emph{IEEE
  transactions on image processing}, vol.~13, no.~4, pp. 600--612, 2004.

\bibitem{KurakinGB16}
\BIBentryALTinterwordspacing
A.~Kurakin, I.~Goodfellow, and S.~Bengio, ``Adversarial examples in the
  physical world,'' 2016. [Online]. Available:
  \url{https://arxiv.org/abs/1607.02533}
\BIBentrySTDinterwordspacing

\end{thebibliography}

\end{document}